\documentclass[]{spie}  

\usepackage{amsmath,amsfonts,amssymb}
\usepackage{graphicx}
\usepackage[colorlinks=true, allcolors=blue]{hyperref}
\usepackage{nicematrix}
\usepackage{multirow}
\usepackage{makecell}
\usepackage[section]{placeins}

\title{SEMI-CenterNet: A Machine Learning Facilitated Approach for 
Semiconductor Defect Inspection}

\author[a, b, *]{Vic De Ridder}
\author[b, *]{Bappaditya Dey}
\author[b,c, *]{Enrique Dehaerne}
\author[b]{Sandip Halder}
\author[b, d]{Stefan De Gendt}
\author[e]{Bartel Van Waeyenberge}
\affil[a]{Faculty of Engineering, Ghent University, Belgium}
\affil[b]{imec, Kapeldreef 75, 3001 Leuven, Belgium}
\affil[c]{Faculty of Science, KU Leuven, 3001 Leuven, Belgium}
\affil[d]{Dept. of Chemistry, KU Leuven, 3001 Leuven, Belgium}
\affil[e]{Dept. of Solid State Sciences, Ghent University, 9000 Ghent, Belgium}
\affil[*]{These authors contributed equally to the work.}
\authorinfo{Further author information: (Send correspondence to both authors)\\Vic De Ridder: E-mail: vic.deridder.ext@imec.be\\Bappaditya Dey: E-mail: bappaditya.dey@imec.be}

\begin{document} 
\maketitle


\begin{abstract}
Continual shrinking of pattern dimensions in the semiconductor domain is making it
increasingly difficult to inspect defects due to factors such as the presence of stochastic noise
and the dynamic behavior of defect patterns and types. Conventional rule-based methods and
non-parametric supervised machine learning algorithms like k-nearest neighbors (kNN) mostly
fail at the requirements of semiconductor defect inspection at these advanced nodes. Deep
Learning (DL)-based methods have gained popularity in the semiconductor defect inspection
domain because they have been proven robust towards these challenging scenarios. In this
research work, we have presented an automated DL-based approach for efficient localization
and classification of defects in SEM images. We have proposed SEMI-CenterNet (SEMI-CN),
a customized CN architecture trained on Scanning Electron Microscope (SEM) images of
semiconductor wafer defects. The use of the proposed CN approach allows improved computational efficiency compared to previously studied DL models. SEMI-CN
gets trained to output the center, class, size, and offset of a defect instance. This is different from
the approach of most object detection models that use anchors for bounding box prediction.
Previous methods predict redundant bounding boxes, most of which are discarded in postprocessing. CN mitigates this by only predicting boxes for likely defect center points. We train SEMI-CN on two datasets and benchmark two ResNet backbones for the framework. Initially, ResNet models pretrained on the COCO dataset undergo training using two datasets separately. Primarily, SEMI-CN shows significant improvement in inference time against previous research works. Finally, transfer learning (using weights of custom SEM dataset) is applied from ADI dataset to AEI dataset and vice-versa, which reduces the required training time for both backbones to reach the best mAP against conventional training method (using COCO dataset pretrained weights).
\end{abstract}


\section{INTRODUCTION}
The continuous shrinking of wafer pattern dimensions is causing increasing difficulties in wafer inspection due to factors such as noise, contrast changes, and lower resolution. SEM has been shown as a useful tool for semiconductor defect inspection due to its high spatial resolution and relatively fast throughput making it suitable for in-line inspection of small defects. Hence SEM imaging is used extensively between different (Litho/Etch) process steps to inspect the patterned wafer. For SEM-based wafer defect inspection, DL-based techniques have been demonstrated as an advantageous technique to deal with the challenging conditions for defect detection caused by shrinking pattern dimensions.\cite{revsemwafer} With continuous developments in Industry 4.0, AI and the Internet of Things, a significant increase in chip demand is inevitable. This may lead to even larger-scale fabs with an exponential increase in the amount of wafer data to be inspected per unit time. To deal with these challenges associated with advanced node defect inspection, fast and lightweight DL models can be beneficial. 

While several works have investigated the use of DL methods for semiconductor defect detection\cite{cnnvsknn, dey2022towards, quantum}, little work has pursued using DL techniques with lower inference time and smaller models. 
Ref.~\citenum{classifautomatic} proposed a lightweight model for automatic classification of semiconductor defects, based on multiple neural networks in a decision tree. While this model achieved good results, it could not localize defects or find multiple defect instances inside one image. 
This work was an initial attempt at developing lightweight DL-based models with subsequently faster inference for semiconductor defect classification and localization. The CenterNet object detection framework\cite{centernet} is investigated and adapted to the semiconductor defect detection task. CenterNet has been demonstrated as a lightweight and fast model family on the object detection task\cite{centernet, centernetb} because it predicts bounding boxes only for probable object centers, which is computationally more efficient compared to anchor-based approaches which predict numerous and redundant bounding boxes in background regions. Additionally, in the current literature,  little investigation has been carried out in applying transfer learning between different semiconductor datasets (ADI-to-AEI and vice-versa) for defect detection application. Since semiconductor wafer SEM images from different datasets share certain features, a model trained from scratch or fine-tuned on one semiconductor SEM dataset can be proven advantageous by sharing weight parameters for extracting and learning subtle local and global features of other SEM semiconductor datasets with numerous defect patterns, compared to models with randomly initialized weights or models trained on other object detection tasks like COCO\cite{coco} or PASCAL-VOC\cite{pascal}. Hence it can be expected that using transfer learning from one semiconductor SEM dataset could lead to a lower number of training epochs on other SEM semiconductor dataset(s) required for satisfactory performance to be achieved because appropriate initial weight parameters help backbones/models to avoid (early) convergence to a suboptimal local minimum of the loss function. In industry, numerous models have already been trained on custom datasets and the transfer learning strategy can emerge as a possible solution for improving Automatic Defect Classification and Detection frameworks.

The main contributions of this research work are i) SEMI-CenterNet, a model designed for fast inference time is proposed for the ADCD task, ii) two ResNet backbone variants integrated into SEMI-CenterNet framework are trained and evaluated for precision and inference time, and compared against previous DL-ADCD models, iii) it is shown that transfer learning between two ADCD datasets can cause shorter training time and comparable performance.

\section{CenterNet object detection framework}
\begin{figure}[h]
    \centering\includegraphics{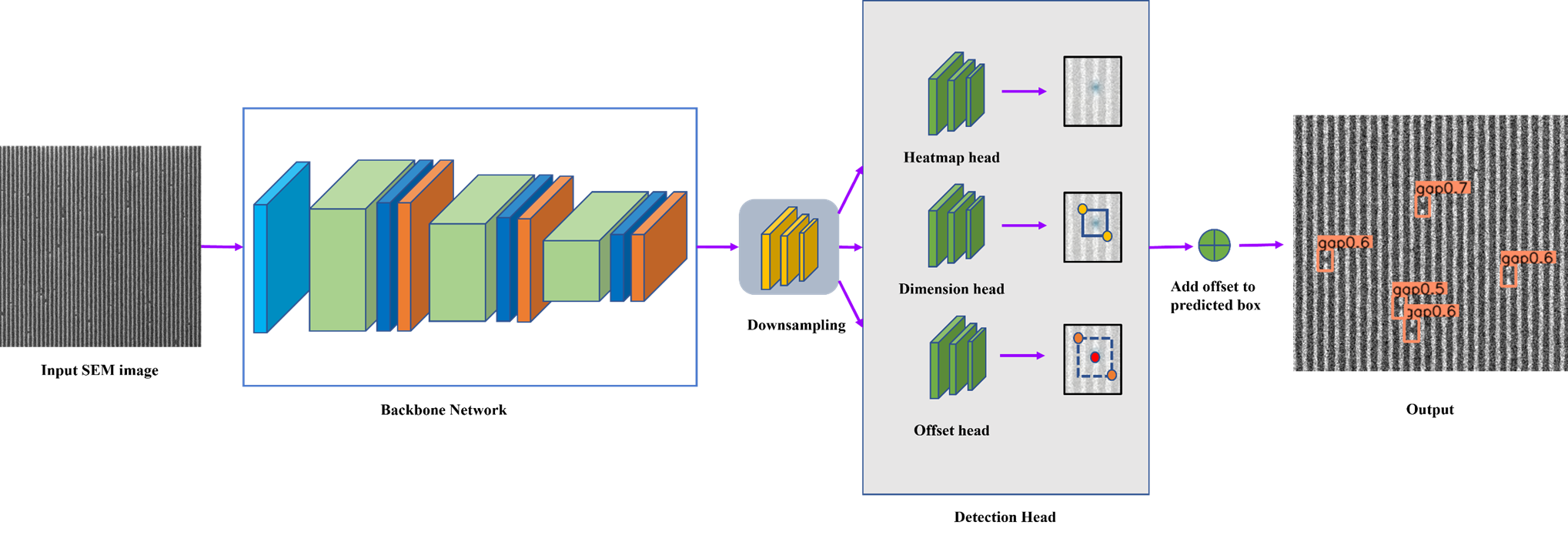}
    \caption{High-level overview of the SEMI-CenterNet based defect detection method adapted from Albahil S. et al.\cite{figure}}
    \label{cnarch}
\end{figure}

 In this section, the CenterNet model is introduced and compared to other object detection architectures. First, DL-based object detection frameworks which were previously studied in the paradigm of semiconductor ADCD are discussed. Afterwards, the conceptual differences between CenterNet and other one-stage/two-stage object detection models are explained and how these differences can lead to inference speed advantages is discussed. Finally, an in-depth look at the CenterNet object detection framework and its underlying mechanisms is provided.
 
Previous works first investigated simple CNN-based defect detection frameworks. In Ref.~\citenum{cnnvsknn}, a simple CNN framework was proposed which generates bounding boxes by classifying sliding window patches as either background or a specific defect type. This work demonstrated better accuracy of CNN-based models compared to other ML-based approaches in semiconductor defect detection. However, the sliding window approach caused a limitation since it creates predetermined bounding box dimensions. This is critical in use cases where large varieties of defects with different sizes exist.

To allow for flexible bounding box sizes and achieve higher accuracy on complex datasets, more complex CNN-based object detection frameworks have been investigated in recent works\cite{yolo7,backbonexamp}. These complex models all use a feature extractor backbone\cite{backbones}. This is a pretrained CNN that extracts relevant high-, mid-, and low-level features from the image. Afterwards, another sequence of CNN-related operations is applied to the extracted feature map to produce the networks' final outputs, often bounding boxes and corresponding class predictions. Due to backbone pretraining, the extracted feature map is more useful for making predictions compared to the original image, resulting in fewer object detection training iterations and often a better final performance compared to models that do not contain a pretrained backbone and are trained for detection from scratch.

The most common, well-studied DL-based object detection frameworks are the Region-based Convolutional Neural Networks (R-CNN)\cite{rcnn} and You Only Look Once (YOLO)\cite{yolo} model families. Both paradigms function through bounding box proposals. A model outputs a large number of bounding boxes and corresponding class confidence scores which are then post-processed using an algorithm such as Non-maximum Suppression (NMS) to discard low confidence predictions and fuse predictions pertaining to the same object. These post-processed predictions are then shown to the user.

Models in the R-CNN family first predict Regions Of Interest (ROIs) through either a CNN or classical approach such as selective search. A bounding box with corresponding confidence scores will then be predicted for each ROI. The YOLO family bypasses the time-intensive ROI generation stage. An image is divided into n same-sized grids (often called anchors) and the model is tasked with predicting k object proposals for each anchor. By using constant ROIs (the grids), YOLO achieves faster inference compared to R-CNN.

The aforementioned object detection model families have been successfully applied to the semiconductor defect detection task as demonstrated in Ref.~\citenum{dey2022towards} and Ref.~\citenum{deyrcnn}. Moreover, architectures are being developed specifically for the task of semiconductor defect detection. In Ref.~\citenum{waferseg}, a model which makes use of a number of regular convolutions in an encoding stage to produce a feature map of the image from which defect segmentations and classifications are predicted using transposed and regular convolutions respectively. However, as of now, no research work has been done to optimize the inference speed of the models.

A downside of the approaches taken by YOLO and R-CNN is, in cases where objects constitute small regions of the input image, a  lot of computations will be involved in predicting bounding boxes for background regions, which is not useful for object detection or localization and thus leads to inefficient use of computational resources. Several works attempt to avoid this issue, among which CenterNet, whose architecture (adapted in the form of SEMI-CN) is shown in figure \ref{cnarch}. CenterNet first predicts probable object centers, and only afterwards constructs bounding boxes for the most likely centers. In this way, computation is first put into detection and center localization and afterward into prediction of object bounds.

To save resources, CenterNet predicts a heatmap of lower resolution than the original image. A stride R is chosen such that for an $n\times n$ pixels image, the model predicts an $\frac{n}{R}\times\frac{n}{R}$ heatmap. To produce the target $\frac{n}{R}\times\frac{n}{R}$ ground truth heatmap, the ground truth center x,y from the $n\times n$ map is scattered on the n/R heatmap around $\frac{x}{R},\frac{y}{R}$ with heat intensity decaying with increasing distance from $\frac{x}{R},\frac{y}{R}$ in a gaussian manner with object size dependent standard deviation.

In model use, bounding boxes eventually need to be displayed for the original image dimensions. The network's heatmap output only consists of discrete points so a discretization error is introduced: the actual ground truth center coordinates in the original image is $c$, the corresponding scaled-down version will be $\frac{c}{R}$ which are not necessarily discrete points. The network makes an offset prediction for each predicted center to resolve this. If a model has predicted center x coordinate $x1$ on the heatmap with corresponding offset $\hat{o}$, the upscaled center x coordinate in the original image becomes $(x1 + \hat{o})*R$. Then the bounding boxes are predicted for each object center using a head separate from the heatmap or offset prediction head, as shown in figure \ref{cnarch}.

\section{Datasets}
\begin{figure}[!h]
    \centering
    \includegraphics[width=0.95\textwidth]{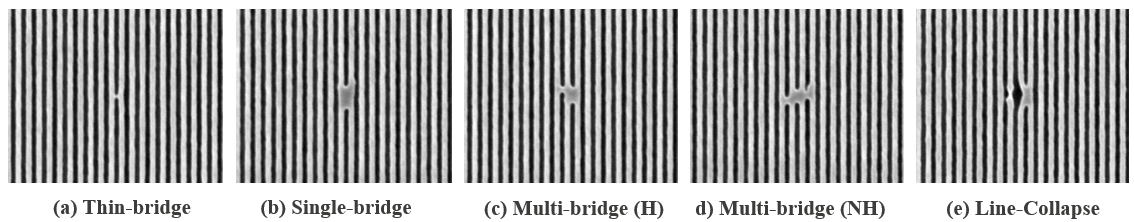}
    \caption{Example defect types in the AEI dataset}
    \label{AEIex}
\end{figure}

\begin{figure}[!h]
    \centering
    \includegraphics[width=0.95\textwidth]{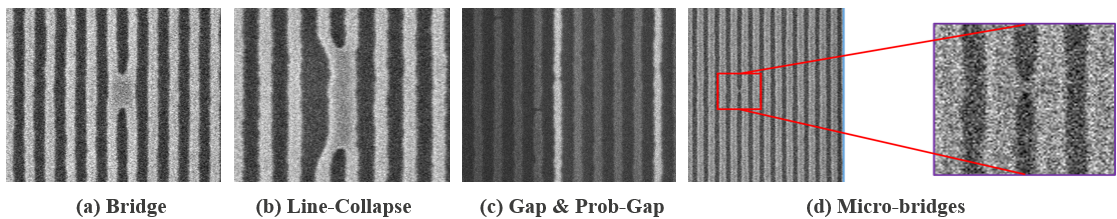}
    \caption{Example defect types in the ADI dataset}
    \label{ADIex}
\end{figure}

The proposed framework is evaluated on two datasets, each containing line-space pattern SEM images taken either After Develop Inspection (ADI dataset) or After Etch Inspection (AEI dataset). Both datasets were already introduced in previous works (Ref.~\citenum{dey2022deep} and Ref.~\citenum{deyrcnn}, respectively). Each dataset was divided into training, validation and test splits as shown in tables \ref{AEI} and \ref{ADI}. Each image is stored in the TIFF greyscale format and each defect has a corresponding bounding box and class annotation generated by a human expert. No synthetic defects or images are used to ensure proper approximation of real fab data and conditions. All defects encountered are stochastic in nature, no intentionally placed defects are present in the datasets. The AEI dataset consists of $480\times480$ pixel images that contain the following defect types: line collapse, single bridge, thin bridge, multi bridge non-horizontal and multi bridge horizontal. Examples of these defects are shown in figure \ref{AEIex}. The ADI dataset consists of $1024\times1024$ pixel images that contain the following defect types: gap, probable gap (p-gap), microbridge, bridge, line-collapse. Examples of these defects are shown in figure \ref{ADIex}.  

\begin{table}[!h]
\centering
\caption{Splits and defect distribution of AEI dataset.}
\label{AEI}
\begin{tabular}{|l|l|l|l|}
\hline
\textbf{Class Name}                                                                        & \textbf{Train (920 images)} & \textbf{Val (120 images)} & \textbf{Test (120 images)} \\ \hline
\textit{\textbf{Linecollapse}}                                                             & 202                         & 34                        & 40                         \\ \hline
\textit{\textbf{Single bridge}}                                                            & 240                         & 31                        & 29                         \\ \hline
\textit{\textbf{Thin bridge}}                                                              & 241                         & 29                        & 29                         \\ \hline
\textit{\textbf{\begin{tabular}[c]{@{}l@{}}Multi bridge \\ (non horizontal)\end{tabular}}} & 160                         & 19                        & 21                         \\ \hline
\textit{\textbf{\begin{tabular}[c]{@{}l@{}}Multi bridge \\ (horizontal)\end{tabular}}}      & 80                          & 10                        & 10                         \\ \hline
\textit{\textbf{Total instances}}                                                          & 923                         & 123                       & 129                        \\ \hline
\end{tabular}
\end{table}

\begin{table}[!h]
\centering
\caption{Splits and defect distribution of ADI dataset.}
\label{ADI}
\begin{tabular}{|l|l|l|l|}
\hline
\textbf{Class Name}               & \textbf{Train (1053 images)} & \textbf{Val (117 images)} & \textbf{Test (154 images)} \\ \hline
\textit{\textbf{gap}}             & 1046                         & 156                       & 174                   \\ \hline
\textit{\textbf{p\_gap}}          & 315                          & 49                        & 54                         \\ \hline
\textit{\textbf{microbridge}}     & 380                          & 47                        & 78                         \\ \hline
\textit{\textbf{bridge}}          & 238                          & 19                        & 17                         \\ \hline
\textit{\textbf{line\_collapse}}  & 550                          & 66                        & 76                         \\ \hline
\textit{\textbf{Total instances}} & 2529                         & 337                       & 399                        \\ \hline
\end{tabular}
\end{table}

\section{Methodology}
\subsection{Benchmarking}\label{bench}
This work will investigate the precision and inference time of the proposed SEMI-CenterNet framework for defect detection on semiconductor wafer images (SEM-based). Our implementation is based on a publicly-available GitHub repository\cite{github}. A performance comparison is made between the use of two different feature extractors/backbones in the proposed framework, ResNet 50 and ResNet 101\cite{ResNet}. Both backbones were trained, initialized with COCO-pretrained weights, on ADI and AEI datasets separately for the semiconductor defect detection task. The inference speed is also recorded for both backbones in terms of required total inference time and pure compute time per image. Per-class Average Precision (AP) and mean Average Precision (mAP) across all defect types are used as performance metrics. The (m)AP results depend on the Intersection over Union (IoU) and confidence thresholds used. The confidence score ranges between 0.0 to 1.0 and is given by the model to express its confidence in the corresponding defect class prediction. By using a confidence threshold, only defect predictions with a confidence score above the threshold are considered when calculating (m)AP. IoU expresses how much the ground truth and predicted bounding box overlap. IoU threshold determines how much this overlap has to be for a prediction to be considered correct. 

For each dataset,  models are trained for 1000 epochs and its performance on the validation dataset is recorded every 50 epochs at confidence threshold 0.33. The per defect class AP and mAP achieved for IoU’s between 0.5 and 0.95 at an interval of 0.05 as well as mAP at an IoU threshold of 0.5 are recorded. The epoch with the best mAP 0.5:0.95 is reported along with the other performance metrics at that epoch.

\subsection{Model transfer}
Another experiment has been conducted (by means of transfer learning), where instead of COCO-pretrained weight initialization, weight parameters corresponding to the highest mAP @IoU0.5:0.95 as described in 4.1 are initialized for finetuning on each dataset. This means the best weight parameters for ADI dataset have been finetuned on AEI dataset and vice versa. This is to validate the possibility of optimizing the model training time (requiring fewer epochs) and other computational resources by applying transfer learning strategy (which means instead of using random weights or weights from completely different data distribution [like COCO], we will initialize weight parameters trained on SEM image features). Hence, the performance is recorded every 20 epochs to monitor this initial behavior. The epoch where the highest mAP IoU0.5:0.95 is achieved on the validation dataset is recorded. Models are finetuned for 500 epochs and the precision and required epochs are compared to the experiments from section \ref{bench}. The best performance achieved by the finetuned model (initialized with custom SEM dataset weight) is compared to that of the models from section \ref{bench} (trained with COCO-pretrained weights).

\section{Results and Discussion}
\begin{figure}[h!]
    \centering
    \includegraphics[width=\textwidth]{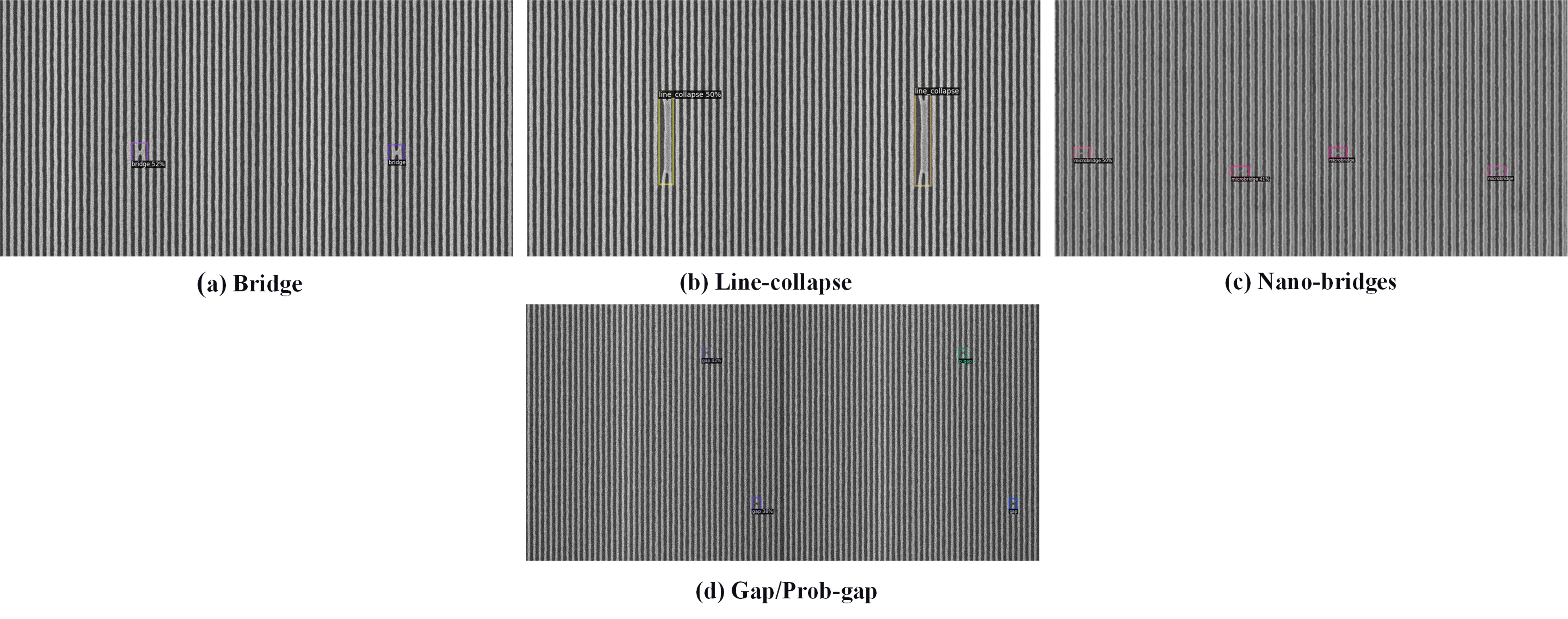}
    \caption{Detection results for each defect type on ADI data. For each pair, (left) model prediction against (right) ground truth.}
    \label{Detection results for each defect type on the ADI dataset}
\end{figure}

\begin{figure}[h!]
    \centering
    \includegraphics[width=\textwidth]{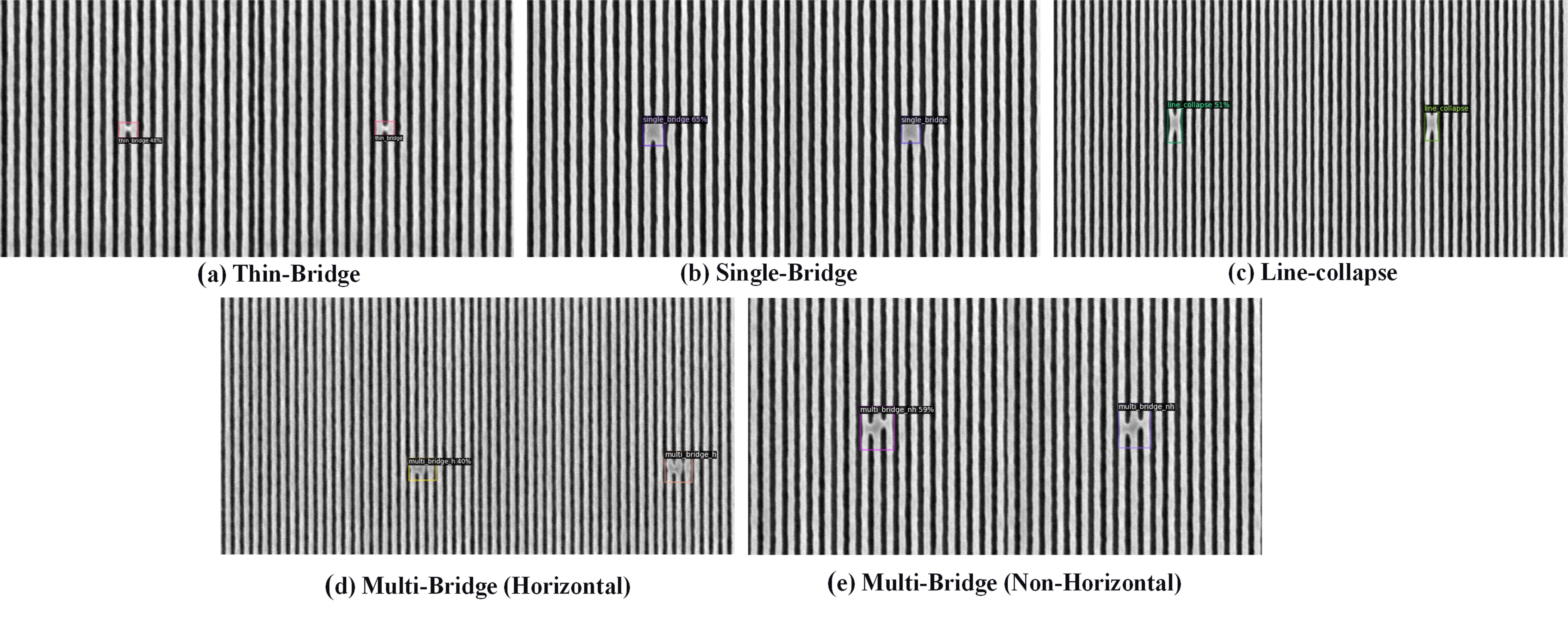}
    \caption{Detection results for each defect type on AEI data. For each pair, (left) model prediction against (right) ground truth.}
    \label{fig:enter-label}
\end{figure}
\subsection{Benchmarking}
\subsubsection{ADI dataset}\label{adires}
The model was evaluated every 50 training epochs and we report the performance of the models for the epoch where the highest validation mAP IoU0.5:0.95 was recorded. Figure \ref{Detection results for each defect type on the ADI dataset} shows detection prediction examples on the ADI dataset. Tables \ref{ADIval} and \ref{ADItest} show the performance on the validation and test datasets respectively. ResNet 50 significantly outperformed ResNet 101 on mAP on both validation and test dataset. However, there is a significant drop (per class AP) for bridge defect for ResNet 50 backbone, from \textbf{35.69} in the validation phase to \textbf{8.78} in the test phase. This drop is a clear outlier compared to other defect types, which requires further investigation.

\begin{table}[h!]
\centering
\caption{Per class and overall average precision on ADI validation data of backbones trained with COCO-pretrained weights with confidence threshold 0.33. Epoch with best mAP IoU0.5:0.95 on validation was selected. Best values in \textbf{bold}.}
\label{ADIval}
\begin{tabular}{|l|l|l|l|l|l|l|l|l|}
\hline
\multirow{2}{*}{\textbf{Backbone}} & \multirow{2}{*}{\textbf{Epoch}} & \multicolumn{5}{|c|}{\textbf{Per class AP IoU0.5:0.95}} & \multirow{2}{*}{\makecell{\textbf{mAP} \\ IoU0.5:0.95}} & \multirow{2}{*}{\makecell{\textbf{mAP} \\ IoU0.5}} \\ \cline{3-7}
 & & microbridge & gap  & bridge & p\_gap & line\_collapse &  &\\ \hline
\textit{\textbf{ResNet 50}} & 750 & \textbf{15.50} & \textbf{31.50} & \textbf{35.69} & \textbf{3.39} & 72.61 & \textbf{31.74} & \textbf{50.97} \\ \hline
\textit{\textbf{ResNet 101}} & 450 & 5.42 & 15.23 & 32.57 & 0.94 & \textbf{74.28} & 25.69 & 42.24 \\ \hline
\end{tabular}
\end{table}

\begin{table}[h!]
\centering
\caption{Per class and overall average precision on ADI test data of backbones trained with COCO-pretrained weights with confidence threshold 0.33. Epoch with best mAP IoU0.5:0.95 on validation was selected. Best values in \textbf{bold}.}
\label{ADItest}
\begin{tabular}{|l|l|l|l|l|l|l|l|l|}
\hline
\multirow{2}{*}{\textbf{Backbone}} & \multirow{2}{*}{\textbf{Epoch}} & \multicolumn{5}{|c|}{\textbf{Per class AP IoU0.5:0.95}} & \multirow{2}{*}{\makecell{\textbf{mAP} \\ IoU0.5:0.95}} & \multirow{2}{*}{\makecell{\textbf{mAP} \\ IoU0.5}} \\ \cline{3-7}
 & & microbridge & gap  & bridge & p\_gap & line\_collapse &  &\\ \hline
\textit{\textbf{ResNet 50}} & 750 & \textbf{14.96} & \textbf{30.50} & 8.78 & \textbf{3.26} & 70.63 & \textbf{25.63} & \textbf{42.77} \\ \hline
\textit{\textbf{ResNet 101}} & 450 & 8.77 & 14.64 & \textbf{24.06} & 1.06 & \textbf{73.29} & 24.36 & 38.34 \\ \hline
\end{tabular}
\end{table}

\subsubsection{AEI dataset}\label{aeires}
Similar to the previous section (\ref{adires}), the model was evaluated every 50 epochs and the best performance on the validation dataset is reported. Figure \ref{fig:enter-label} shows detection prediction examples on the AEI dataset.  Tables \ref{aeival} and \ref{AEItest} show the performance on the validation and test dataset respectively. ResNet 101 significantly outperforms ResNet 50. While the performance difference between the two backbones remains significant on both test and validation, the per class AP by ResNet 101 on the Multi Bridge Horizontal (MBH) defect type drops from \textbf{34.85} on validation to \textbf{4.36} on test. Similar to section \ref{adires}, the reasoning behind such a large drop can be investigated in future work. Additionally, ResNet 50 fails to detect any Multi Bridge Non Horizontal (MBNH) defects and achieves extremely small AP values of \textbf{4.32} and \textbf{6.95} on thin bridge defect type. Stochastic failure analysis on certain defect types may be done in future work.

\begin{table}[h!]
\centering
\caption{Per class and overall average precision on AEI validation data of backbones trained with COCO-pretrained weights with confidence threshold 0.33. Epoch with best mAP IoU0.5:0.95 on validation was selected. Best values in \textbf{bold}.}
\label{aeival}
\begin{tabular}{|l|l|l|l|l|l|l|l|l|}
\hline
\multirow{2}{*}{\textbf{Backbone}} & \multirow{2}{*}{\textbf{Epoch}} & \multicolumn{5}{|c|}{\textbf{Per class AP IoU0.5:0.95}} & \multirow{2}{*}{\makecell{\textbf{mAP} \\ @0.5:0.95}} & \multirow{2}{*}{\makecell{\textbf{mAP} \\ @0.5}} \\ \cline{3-7}
 & & thin\_bridge & single\_bridge  & MBNH & MBH & line\_collapse &  &\\ \hline
\textit{\textbf{ResNet50}} & 200 & 4.32 & 15.65 & 0.00 & 26.48 & \textbf{65.84} & 22.46 & 39.46 \\ \hline
\textit{\textbf{ResNet101}} & 450 & \textbf{26.99} & \textbf{38.78} & \textbf{28.59} & \textbf{34.85} & 40.406 & \textbf{33.92} & \textbf{53.18} \\ \hline
\end{tabular}
\end{table}

\begin{table}[h!]
\centering
\caption{Per class and overall average precision on AEI test data of backbones trained with COCO-pretrained weights with confidence threshold 0.33. Epoch with best mAP IoU0.5:0.95 on validation was selected. Best values in \textbf{bold}.}
\label{AEItest}
\begin{tabular}{|l|l|l|l|l|l|l|l|l|}
\hline
\multirow{2}{*}{\textbf{Backbone}} & \multirow{2}{*}{\textbf{Epoch}} & \multicolumn{5}{|c|}{\textbf{Per class AP IoU0.5:0.95}} & \multirow{2}{*}{\makecell{\textbf{mAP} \\ @0.5:0.95}} & \multirow{2}{*}{\makecell{\textbf{mAP} \\ @0.5}} \\ \cline{3-7}
 & & thin\_bridge & single\_bridge  & MBNH & MBH & line\_collapse &  &\\ \hline
\textit{\textbf{ResNet50}} & 200 & 6.95 & 32.60 & 0.00 & \textbf{20.17} & \textbf{57.75} & 23.49 & 39.14 \\ \hline
\textit{\textbf{ResNet101}} & 450 & \textbf{24.67} & \textbf{42.77} & \textbf{32.22} & 4.36 & 45.23 & \textbf{29.85} & \textbf{54.25} \\ \hline
\end{tabular}
\end{table}

\subsection{Model Transfer}
Tables \ref{aeitoadi} and \ref{aditoaei} show the comparison in validation precision for models trained in sections \ref{adires} and \ref{aeires} against models that have been fine-tuned with weight parameter initialization, trained on SEM image features. On the ADI dataset, finetuning the model from the AEI dataset pretrained weight shows a significant decrease in the number of required epochs before the best precision on validation data is reached. Retraining the ResNet 50-based model required only 80 epochs instead of 750 epochs before it reached its highest precision for the transfer learning strategy. For ResNet 101, 100 epochs instead of 450 epochs of training before convergence was observed as well as an improvement in detection mAP. On the AEI dataset, both significantly outperform against backbones trained with COCO-pretrained weight parameter. ResNet 50 and ResNet 101 (finetuned with ADI dataset-pretrained weight parameter) improve AEI test mAP @IoU0.5:0.95 by \textbf{37.42\%} and \textbf{19.50\%}, against models trained with COCO-pretrained weight parameter. In general observation, all models trained/fine-tuned with SEM dataset pretrained weights converge faster than models initialized with random/COCO-pretrained weights, especially on ADI dataset training. In general, AEI dataset is less complex compared to the ADI dataset. Most images have only one defect instance per image and attribution of noise pixels and contrast change scenario is less significant. Hence our experimental observations indicate that weights pretrained on semiconductor image datasets can be advantageous for overall training from scratch and/or fine-tuning as well as when mainly using a model pretrained on complex semiconductor data and finetuning it for less complex data.

\begin{table}[h!]
\centering
\caption{mAP scores and epoch needed on ADI dataset for backbones trained with COCO-pretrained weights against backbones trained with AEI dataset-pretrained weights. Best metric achieved per backbone in \textbf{bold}.}
\label{aeitoadi}
\begin{tabular}{|l|l|l|l|l|l|l|l|}
\hline
\multirow{2}{*}{\textbf{Backbone}} & \multicolumn{3}{|c|}{\textbf{Normal initialization}} & \multicolumn{4}{|c|}{\textbf{Finetuning from AEI}} \\ \cline{2-8}
& Epoch & \makecell{Val mAP \\ IoU0.5:0.95} & \makecell{Test mAP\\ IoU0.5:0.95} &Epoch & \makecell{Val mAP\\ IoU0.5:0.95} & \makecell{Test mAP \\IoU0.5:0.95}  & \makecell{Test mAP \\ IoU0.5}\\ \hline
\textit{\textbf{ResNet 50}} & 750  & \textbf{31.74} & \textbf{25.63} & \textbf{80} & 22.92 & 23.23 & 38.64\\ \hline
\textit{\textbf{ResNet 101}} & 450 & 25.69 & 24.63 & \textbf{100} & \textbf{28.43} & \textbf{28.28} & 47.89\\ \hline
\end{tabular}
\end{table}

\begin{table}[h!]
\centering
\caption{mAP scores and epoch needed on AEI dataset for backbones trained with COCO-pretrained weights against backbones trained with AEI dataset-pretrained weights. Best metric achieved per backbone in \textbf{bold}.}
\label{aditoaei}
\begin{tabular}{|l|l|l|l|l|l|l|l|}
\hline
\multirow{2}{*}{\textbf{Backbone}} & \multicolumn{3}{|c|}{\textbf{Normal initialization}} & \multicolumn{4}{|c|}{\textbf{Finetuning from ADI}} \\ \cline{2-8}
& Epoch & \makecell{Val mAP \\ IoU0.5:0.95} & \makecell{Test mAP\\ IoU0.5:0.95} &Epoch & \makecell{Val mAP\\ IoU0.5:0.95} & \makecell{Test mAP \\IoU0.5:0.95} & \makecell{Test mAP \\ IoU0.5}\\ \hline
\textit{\textbf{ResNet50}} & 200  & 22.46 & 23.49 & \textbf{160} & \textbf{34.68} & \textbf{32.28} & 61.04\\ \hline
\textit{\textbf{ResNet101}} & 450 & \textbf{33.92}& 29.85&\textbf{440} & 31.23 & \textbf{35.67} & 61.47\\ \hline
\end{tabular}
\end{table}

\subsection{Comparison to previous work}
Table \ref{compadi} shows the inference speed and mAPs achieved on the ADI dataset by the proposed SEMI-CenterNet and previously studied models from Ref.~\citenum{revsemwafer}. mAPs are all reported on the test data split. While the mAP achieved by SEMI-CenterNet-based models is worse compared to previous work, it does achieve the lowest inference time yet. Compared to the second fastest model (YOLOv7), SEMI-CenterNet with ResNet 50 backbone achieves \textbf{55.94\%} improvement in inference time per image. This demonstrates models based on the CenterNet architecture are capable of faster inference and show potential, but further investigation on lower precision is required before it can be proposed as a valid alternative to models from previous works\cite{yolo7, dey2022deep, deyrcnn}.

\begin{table}[h]
\centering
\caption{Inference time and mAP performance comparison between different models on ADI dataset. Best values in \textbf{bold}.}
\label{compadi}
\begin{tabular}{|l|l|l|}
\hline
\textbf{Model} & \textbf{mAP IoU0.5} & \textbf{Inference time (ms / image)} \\ \hline
\textit{\textbf{\begin{tabular}[c]{@{}l@{}}Faster R-CNN\\ Faster R-CNN (+NWD)\end{tabular}}} & \begin{tabular}[c]{@{}l@{}}0.825\\ 0.827\end{tabular} & \begin{tabular}[c]{@{}l@{}}56.8\\ 60.2\end{tabular} \\ \hline
\textit{\textbf{\begin{tabular}[c]{@{}l@{}}DINO (ResNet-50)\\ DINO (SWIN-Tiny)\end{tabular}}} & \begin{tabular}[c]{@{}l@{}}\textbf{0.865}\\ 0.769\end{tabular} & \begin{tabular}[c]{@{}l@{}}108.7\\ 119.2\end{tabular} \\ \hline
\textbf{RetinaNet (ResNet-152) \cite{dey2022deep}} & 0.788 & 78.2 \\ \hline
\textbf{YOLOv7 \cite{yolo7}} & 0.843 & 20.2 \\ \hline
\textbf{\begin{tabular}[c]{@{}l@{}}Proposed SEMI-CenterNet \\ (ResNet-50)\end{tabular}} & 0.472 & \begin{tabular}[c]{@{}l@{}}\textbf{8.9}\\ 8.7 (pure compute time)\end{tabular} \\ \hline
\textbf{\begin{tabular}[c]{@{}l@{}}Proposed SEMI-CenterNet\\ (ResNet-101)\end{tabular}} & 0.479 & \begin{tabular}[c]{@{}l@{}}13.4\\ 12.8 (pure compute time)\end{tabular} \\ \hline
\end{tabular}
\end{table}

\section{Conclusion}
In this work, the SEMI-CenterNet framework was proposed to reduce the computation time for DL-based semiconductor Automated Defect Classification and Detection. Due to its anchorless architecture, the model makes fewer predictions on background image sections, leading to faster inference. ResNet 50 and 101 backbones were investigated on both an ADI and AEI dataset. Despite its smaller size, ResNet 50 performed best on the ADI dataset, while ResNet 101 showed better performance on AEI dataset. Moreover, it was shown that using the model weights obtained by training on one semiconductor wafer dataset to retrain/fine-tune and retraining them on other wafer datasets (ADI/AEI) causes models to converge to similar performance faster since the model had learned various semiconductor defect pattern features and distributions. Finally, we compared our proposed framework against previous research works on the same ADI dataset, both on precision and inference time. Future work can be extended toward analyzing the deteriorated performance of certain classes in certain situations to improve detection precision metrics.


\bibliography{report} 

\begin{thebibliography}{10}

\bibitem{revsemwafer}
Dehaerne, E., Dey, B., and Halder, S., ``A comparative study of deep-learning
  object detectors for semiconductor defect detection,'' in [{\em 2022 29th
  IEEE International Conference on Electronics, Circuits and Systems
  (ICECS)}{\nolinebreak\hspace{0.1em}]},   1--2, IEEE (2022).

\bibitem{cnnvsknn}
Cheon, S., Lee, H., Kim, C.~O., and Lee, S.~H., ``Convolutional neural network
  for wafer surface defect classification and the detection of unknown defect
  class,'' {\em IEEE Transactions on Semiconductor Manufacturing}~{\bf 32}(2),
  163--170 (2019).

\bibitem{dey2022towards}
Dey, B., Dehaerne, E., and Halder, S., ``Towards improving challenging
  stochastic defect detection in sem images based on improved yolov5,'' in
  [{\em Photomask Technology 2022}{\nolinebreak\hspace{0.1em}]},   {\bf 12293},
   28--37, SPIE (2022).

\bibitem{quantum}
Yang, Y.-F. and Sun, M., ``Semiconductor defect detection by hybrid
  classical-quantum deep learning,'' in [{\em 2022 {IEEE}/{CVF} Conference on
  Computer Vision and Pattern Recognition
  ({CVPR})}{\nolinebreak\hspace{0.1em}]},  {IEEE} (jun 2022).

\bibitem{classifautomatic}
Li, Z., Wang, Z., and Shi, W., ``Automatic wafer defect classification based on
  decision tree of deep neural network,'' in [{\em 2022 33rd Annual SEMI
  Advanced Semiconductor Manufacturing Conference
  (ASMC)}{\nolinebreak\hspace{0.1em}]},   1--6, IEEE (2022).

\bibitem{centernet}
Zhou, X., Wang, D., and Kr{\"a}henb{\"u}hl, P., ``Objects as points,'' {\em
  arXiv preprint arXiv:1904.07850}  (2019).

\bibitem{centernetb}
Duan, K., Bai, S., Xie, L., Qi, H., Huang, Q., and Tian, Q., ``Centernet:
  Keypoint triplets for object detection,'' in [{\em Proceedings of the
  IEEE/CVF international conference on computer
  vision}{\nolinebreak\hspace{0.1em}]},   6569--6578 (2019).

\bibitem{coco}
Lin, T.-Y., Maire, M., Belongie, S., Hays, J., Perona, P., Ramanan, D.,
  Doll{\'a}r, P., and Zitnick, C.~L., ``Microsoft coco: Common objects in
  context,'' in [{\em Computer Vision--ECCV 2014: 13th European Conference,
  Zurich, Switzerland, September 6-12, 2014, Proceedings, Part V
  13}{\nolinebreak\hspace{0.1em}]},   740--755, Springer (2014).

\bibitem{pascal}
Hoiem, D., Divvala, S.~K., and Hays, J.~H., ``Pascal voc 2008 challenge,'' {\em
  World Literature Today}~{\bf 24}(1) (2009).

\bibitem{figure}
Albahli, S. and Nazir, T., ``Ai-centernet cxr: An artificial intelligence (ai)
  enabled system for localization and classification of chest x-ray disease,''
  {\em Frontiers in Medicine}~{\bf 9},  955765 (2022).

\bibitem{yolo7}
Dehaerne, E., Dey, B., Halder, S., and De~Gendt, S., ``Optimizing yolov7 for
  semiconductor defect detection,'' in [{\em Metrology, Inspection, and Process
  Control XXXVII}{\nolinebreak\hspace{0.1em}]},   {\bf 12496},  635--642, SPIE
  (2023).

\bibitem{backbonexamp}
Kim, J., Nam, Y., Kang, M.-C., Kim, K., Hong, J., Lee, S., and Kim, D.-N.,
  ``Adversarial defect detection in semiconductor manufacturing process,'' {\em
  IEEE Transactions on Semiconductor Manufacturing}~{\bf 34}(3),  365--371
  (2021).

\bibitem{backbones}
Elharrouss, O., Akbari, Y., Almaadeed, N., and Al-Maadeed, S.,
  ``Backbones-review: Feature extraction networks for deep learning and deep
  reinforcement learning approaches,'' {\em arXiv preprint arXiv:2206.08016}
  (2022).

\bibitem{rcnn}
Ren, S., He, K., Girshick, R., and Sun, J., ``Faster r-cnn: Towards real-time
  object detection with region proposal networks,'' {\em Advances in neural
  information processing systems}~{\bf 28} (2015).

\bibitem{yolo}
Redmon, J., Divvala, S., Girshick, R., and Farhadi, A., ``You only look once:
  Unified, real-time object detection,'' in [{\em Proceedings of the IEEE
  conference on computer vision and pattern
  recognition}{\nolinebreak\hspace{0.1em}]},   779--788 (2016).

\bibitem{deyrcnn}
Dey, B., Dehaerne, E., Halder, S., Leray, P., and Bayoumi, M.~A., ``{Deep
  learning based defect classification and detection in SEM images: a mask
  R-CNN approach},'' in [{\em Metrology, Inspection, and Process Control
  XXXVI}{\nolinebreak\hspace{0.1em}]},   {\bf PC12053},  PC120530K, SPIE
  (2022).

\bibitem{waferseg}
Nag, S., Makwana, D., Mittal, S., Mohan, C.~K., et~al., ``Wafersegclassnet-a
  light-weight network for classification and segmentation of semiconductor
  wafer defects,'' {\em Computers in Industry}~{\bf 142},  103720 (2022).

\bibitem{dey2022deep}
Dey, B., Goswami, D., Halder, S., Khalil, K., Leray, P., and Bayoumi, M.~A.,
  ``Deep learning-based defect classification and detection in sem images,'' in
  [{\em Metrology, Inspection, and Process Control
  XXXVI}{\nolinebreak\hspace{0.1em}]},   PC120530Y, SPIE (2022).

\bibitem{github}
Wang, F., ``Centernet-better.''
  \url{https://github.com/FateScript/CenterNet-better} (2020).

\bibitem{ResNet}
He, K., Zhang, X., Ren, S., and Sun, J., ``Deep residual learning for image
  recognition,'' in [{\em Proceedings of the IEEE conference on computer vision
  and pattern recognition}{\nolinebreak\hspace{0.1em}]},   770--778 (2016).

\end{thebibliography}
\bibliographystyle{spiebib}

\end{document}